\pdfoutput=1

\documentclass[11pt]{article}

\usepackage{acl}

\usepackage{times}
\usepackage{latexsym}

\usepackage[T1]{fontenc}

\usepackage[utf8]{inputenc}

\usepackage{microtype}

\usepackage{inconsolata}

\usepackage{subcaption}
\usepackage{amsmath}
\usepackage{amssymb}
\usepackage{bm}
\usepackage{algorithm}
\usepackage[noend]{algpseudocode}
\usepackage{multirow}
\usepackage{booktabs} 
\usepackage{tabularx}
\usepackage{xcolor}
\usepackage{listings,multicol}
\usepackage{graphicx} 
\usepackage{rotating}
\usepackage{transparent}
\usepackage{soul}
\usepackage{mdframed}

\definecolor{lightred}{RGB}{196, 73, 72}
\definecolor{lightblue}{RGB}{ 73, 72, 196}
\definecolor{pink}{RGB}{183, 178, 254}
\definecolor{cyan}{RGB}{205, 207, 196}
\definecolor{codegreen}{rgb}{0,0.6,0}
\definecolor{codegray}{rgb}{0.1,0.1,0.1}
\definecolor{codepurple}{rgb}{0.58,0,0.82}
\definecolor{backcolour}{rgb}{1.0,1.0,1.0}

\title{Controllable Citation Sentence Generation with Language Models}

\author{Nianlong Gu \and Richard H.R. Hahnloser \\
        Institute of Neuroinformatics, \\
        University of Zurich and ETH Zurich\\
        \texttt{\{nianlong,rich\}@ini.ethz.ch}
        }

\begin{document}
\maketitle
\begin{abstract}

Citation generation aims to generate a citation sentence that refers to a chosen paper in the context of a manuscript. However, a rigid citation generation process is at odds with an author's desire to control specific attributes, such as 1) the citation intent, e.g., either introducing background information or comparing results, and 2) keywords that should appear in the citation text. To provide these degrees of controllability during citation generation, we propose to integrate the manuscript context, the context of the referenced paper, and the desired control attributes into a structured template and use it to fine-tune a language model (LM) via next-token prediction. We then utilize Proximal Policy Optimization to directly optimize the LM in favor of a high score of our proposed controllability metric. The proposed workflow harmoniously combines citation attribute suggestion and conditional citation generation into one LM, allowing for better user control.\footnote{
Our code and data are available at \url{https://github.com/nianlonggu/LMCiteGen}
} 

\end{abstract}

\section{Introduction}
\label{sec:intro}

A common practice in scientific writing is to cite and discuss relevant papers that support an argument, provide background information, or compare results \cite{2018tenrules}. Recent studies aim to facilitate this citation process by using neural networks to generate a citation sentence based on the context of the manuscript and the paper to be cited.  \citet{nikiforovskaya_automatic_2020} proposed a BERT-based extractive summarizer \cite{Liu2019} that produces a paper review by extracting one sentence from each of the related papers. \citet{chen_automatic_2019} proposed automatically generating a related work section by extracting information on how papers in the reference list have been cited in previous articles. \citet{xing_automatic_2020} developed an RNN-based pointer generator network that can copy words from the manuscript and the abstract of the cited paper based on cross-attention. \citet{ge_baco_2021} further extended this work by enhancing citation generation using information from the citation graph.

These efforts focused primarily on developing a sequence-to-sequence pipeline that works in a fully automated, uncontrolled manner, leaving little room for users to control the generation process. 
We believe control is desirable because authors often have a clear motivation before writing a citation sentence. For example, they may have a specific \textit{intent} to cite, such as comparing results or presenting background information, or they may have \textit{keywords} in mind to appear in the citation sentence.
When the generated citation does not match an author's motivation, the author may wish to change the generation by specifying certain attributes, such as citation intent or keywords. 



This study aims to develop a citation generation model that is controllable, such that users can alter the citation intent or topic by explicitly providing the citation intent and keywords.
Our proposed method involves the following two phases: 

\noindent\textbf{Supervised fine-tuning}. We design a structured prompt template that systematically and consecutively incorporates contextual information, citation attributes (intent and keywords), and the citation sentence into a sequence of tokens, and fine-tune an LM via next-token prediction. Through supervised fine-tuning, the LM learns to generate citation sentences not only based on the manuscript/cited paper's context but also conditioned on the citation attributes, thus allowing flexible control of generation by altering the citation attributes \cite{https://doi.org/10.48550/arxiv.1909.05858}.

\noindent\textbf{Controllability enhancement via reinforcement learning}. We propose measuring the controllability of a citation generation system from multiple aspects with the following metrics: i. Intent Alignment Score (IAS), which measures whether the intent of the generated citation sentence matches the given control intent; ii. Keyword Recall (KR), which measures the recall of the control keywords in the generated citation; iii. Fluency Score (FS); and iv. ROUGE-F1 \cite{lin-2004-rouge} score of the generated sentence compared with the groud-truth citation.
These controllability evaluation metrics provide a further opportunity to guide the training of our system by using them to estimate a reward function for Proximal Policy Optimization (PPO) \cite{schulman2017proximal}. This allows us to explore the effect of using feedback from our chosen metrics to improve our system's controllability.




Our contributions are summarized as follows: 
\begin{itemize}
    \item We present a novel strategy that unifies citation attributes and citation sentence generation within one language model, thereby enabling user control of the citation generation process.
    \item We evaluate the control exerted by the various attributes, employing a multi-metric system that includes an intent matching score, keyword recall, ROUGE score, and a reference-less fluency score, and we use these controllability metrics as a reward for Proximal Policy Optimization (PPO) to effectively improve the controllability of our model following its initial supervised training.
    \item  We curate a comprehensive dataset, parsing both contextual text and citation attributes, to offer a valuable resource for future controllable citation generation research.
\end{itemize}

\section{Related Work}

Previous work in citation generation, including \citet{xing_automatic_2020,ge_baco_2021,Wang_Song_Li_Cheng_Ju_Zhang_Wang_2022}, approached the task as a sequence-to-sequence translation problem, utilizing recurrent networks and knowledge graph enhancements.
Despite their advances, these approaches do not fully address the complexities of citation sentence generation. Our research highlights the importance of user-specified attributes and illustrates the limitations of large language models in reliably inferring key attributes such as topic keywords (Table \ref{tab:res_attr_matching}). We propose that this semantic gap necessitates a shift towards models that take explicit user control into account in citation generation.

Recent research \citet{jung2022intent, https://doi.org/10.48550/arxiv.2112.01332,yu2022scientific}  has also explored controlled citation generation, though the focus is often limited to controlling a single attribute like citation intent. Our work extends this concept to a broader range of attributes, introducing methods to suggest potential attributes and balance the automation and controllability of citation generation. We further differentiate our research by conducting extensive experiments with cutting-edge language models and investigating the augmentation of controllability via reinforcement learning, contributing to a more comprehensive understanding of controlled citation generation. In addition, we go beyond prompt-based approaches \cite{yang2022tailor} by investigating enhancing the controllability of our citation generation system with reinforcement learning.

\section{Method}
\begin{figure*}
\centering
  \includegraphics[width=\linewidth]{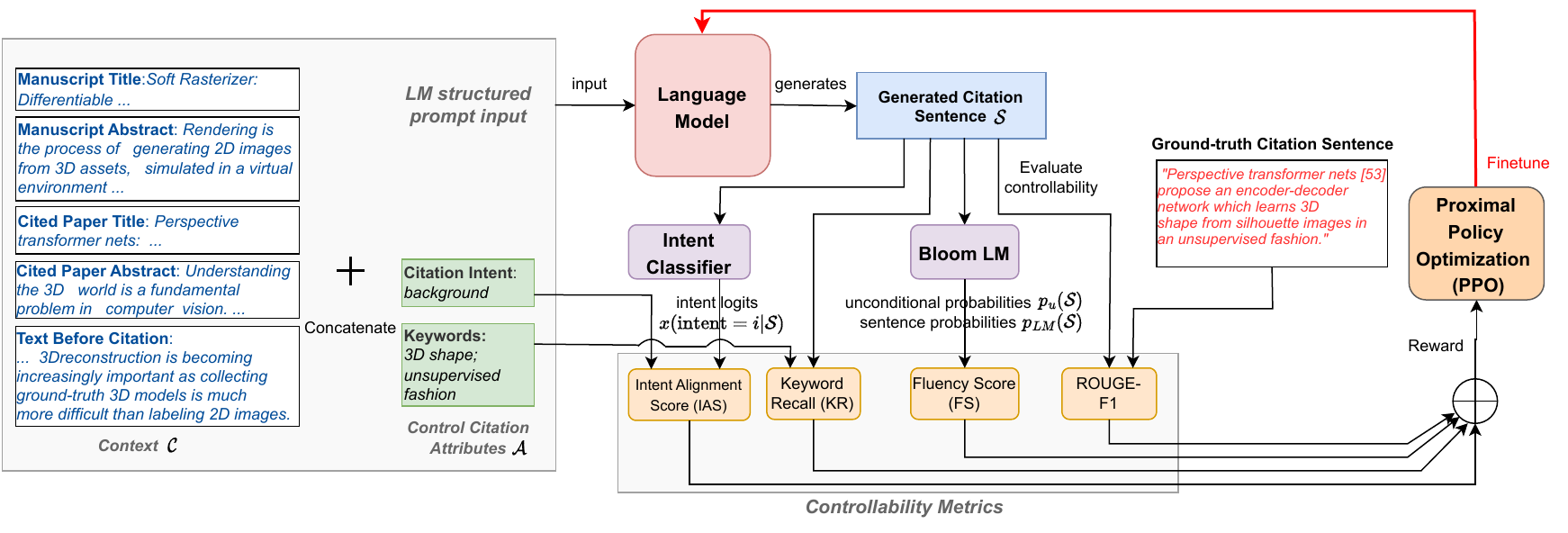}
  \caption{
A schematic representation of the workflow for generating citation sentences using a Language Model, with subsequent controllability evaluation, and optimization using Proximal Policy Optimization (PPO) with the summed controllability metrics as a reward.
  }
  \label{fig:overall-pipeline}
\end{figure*}
We first describe our supervised method for fine-tuning an LM, and then we describe how we measure the control of the generated sentence by the attributes. Finally, we introduce strategies for enhancing controllability through reinforcement learning. Our method is illustrated in Figure \ref{fig:overall-pipeline}.

\subsection{LM Supervised Fine-tuning}
\label{sec:attribute_gen}
We fine-tune a language model (LM) to generate a citation sentence $\mathcal{S}$ given the context $\mathcal{C}$ and citation (control) attributes $\mathcal{A}$. The context, denoted as $\mathcal{C}$, consists of information from three distinct sources: 1) the local manuscript context, including up to $N_s$ sentences from the same section (we use $N_s=5$) that precede the citation sentence to be generated; 2) the manuscript's title and abstract, serving as the global context; and 3) the title and abstract of the paper to be cited, providing an external context.

The citation attributes denoted $\mathcal{A}$, include the combined citation intent and keywords. Following the framework proposed in \citet{cohan-etal-2019-structural}, we define three categories of citation intents: 
1) \textit{background}: The citation offers context or background information about a pertinent problem, concept, method, or topic.
2) \textit{method}: The citation refers to a specific method, tool, approach, or dataset in the cited paper.
3) \textit{result}: The citation contrasts or compares the results or findings of the manuscript with those in the referenced paper.
In terms of keywords, we define keyword attributes as one or two noun phrases extracted from the target citation sentence that bears semantic similarity to the context of either the manuscript or the cited paper. 

The training objective is to optimize the LM to maximize the log-likelihood of generating both the citation attributes $\mathcal{A}$ and the citation sentence $\mathcal{S}$ given the context $\mathcal{C}$. 
Let \( a_i \) and \( s_i \) be the tokens in  \( \mathcal{A} \) and \( \mathcal{S} \), with total tokens \( |A| \) and \( |S| \) respectively, the training objective can be expressed as:
\begin{equation}
\begin{aligned}
     \hat{\theta} &= \arg \max_{\theta}  \log p(\mathcal{S}, \mathcal{A}|\mathcal{C}) \\
     & = \arg \max_{\theta} \left[ \log p(\mathcal{A}|\mathcal{C})  + \log p(\mathcal{S}|\mathcal{A},\mathcal{C}) \right]\\
    &=  \arg \max_{\theta} \Bigg[ \sum_{i=1}^{|\mathcal{A}|}\log p( a_i|a_1\dots a_{i-1}, \mathcal{C} )\ + \\ & \ \ \ \ \ \ \ \ \ \ \ \ \ \ \ \ 
\ \ \sum_{i=1}^{|\mathcal{S}|} \log p(s_i|s_1\dots s_{i-1},\mathcal{A},\mathcal{C}) \Bigg]
\end{aligned}
\label{eq:joint-prob}
\end{equation}

Such an objective allows us to fine-tune a single LM for two purposes: 1) inferring citation attributes \( \mathcal{A} \) given the context \( \mathcal{C} \), and 2) crafting citation sentences \( \mathcal{S} \) based on the context \( \mathcal{C} \) and control attributes \( \mathcal{A} \). This dual-purpose training strategy enables the fine-tuned LM to switch dynamically between controlled and uncontrolled modes: In the \textbf{controlled mode}, the LM assimilates the context and the user-specified control attributes to generate the citation sentence. In the \textbf{uncontrolled mode}, the LM initiates by automatically inferring possible citation attributes based solely on the context, and uses these inferred attributes to guide itself when generating the citation sentence. This flexibility also allows us to compare the performance of our citation generation model between uncontrolled and controlled modes in a fair manner, since we can compare the performance on the same model just with different working modes.

\textbf{Structured Input Template for Supervised Fine-tuning}.
We design an input template to help the LM differentiate between input sources. The template arranges different input components into a unique prompt, including the cited paper's global context, the manuscript's local and global contexts, and the citation attributes, as in \citet{JMLR:v21:20-074,https://doi.org/10.48550/arxiv.1909.05858,alpaca}. The final input text is structured as follows:
\begin{mdframed}[linewidth=1pt]
\texttt{\#\#\#Manuscript Title: XXX XXX}

\noindent\texttt{\#\#\#Manuscript Abstract: XXX XXX}

\noindent\texttt{\#\#\#Cited Paper Title: XXX XXX} 

\noindent\texttt{\#\#\#Cited Paper Abstract: XXX XXX} 

\noindent\texttt{\#\#\#Text Before Citation:} 

\noindent$\ \ \ \ \ \ \ \ \ \ \ \ \ \ \ \ \ \ $\texttt{XXX XXX} ({\small manuscript local context})

\noindent\texttt{\#\#\#Citation Intent: XXX} 
({\small one word from `background', `method' and `result'})

\noindent\texttt{\#\#\#Keywords: XXX; XXX} ({\small keywords relevant to the citation, separated by `; '})

\noindent\texttt{\#\#\#Citation: XXX XXX} ({\small target citation sentence})
\end{mdframed}


For decoder-only LMs like GPT-NEO \cite{gpt-neo}, the fine-tuning phase uses the generated prompt as input for next-token prediction, masking context-related tokens (from ``{\texttt{\#\#\#Manuscript Title: XXX}}'' to ``{\texttt{\#\#\#Text Before Citation: XXX}}'') to confine prediction loss within citation attributes $\mathcal{A}$ and citation sentence $\mathcal{S}$, as per Equation \eqref{eq:joint-prob}. In the inference stage, the LM is fed with the template up to ``{\texttt{\#\#\#Citation Intent:}}'', ``{\texttt{\#\#\#Keywords:}}'', and ``{\texttt{\#\#\#Citation:}}'' when the task is to decode the citation intent, keywords, and the citation sentence, respectively.
In contrast, for encoder-decoder LMs like BART \cite{lewis-etal-2020-bart}, context-related tokens are the encoder input during both fine-tuning and inference. During fine-tuning, the LM learns to decode tokens within $\mathcal{A}$ and $\mathcal{S}$. At inference, the decoder generates citation attributes and sentences based on the provided prompt.

\subsection{ Controllability Evaluation Metrics}
\label{sec:control-eval}
We deem good controllability by control attributes can be reflected in the following aspects:
\begin{enumerate}
    \item The generated citation sentence matches the given intent. For example, given a control intent ``background'', the model will generate a sentence that introduces the background of the cited paper.
    \item The generated citation sentence contains the given control keywords.
    \item The generated sentence is fluent so that the control keywords are embedded into the sentence in a logically coherent way.
    \item The generated sentence is content-wise related to the cited paper and fits well with the context of the manuscript.
\end{enumerate}

Along this vein, apart from the human evaluation (Section \ref{sec:human-eval}), we propose the following four automatic metrics, each corresponding to one aspect:

1. \textbf{Intent Alignment Score (IAS)} evaluates the alignment between the generated citation sentence and the specified citation intent. Suppose the control intent is $i$ (one of three possible intents: `\textit{background}', `\textit{method}', and `\textit{result}') and LM generates a citation sentence $\mathcal{S}$,
we use SciBERT \cite{beltagy-etal-2019-scibert} to process the citation sentence (preceded by a ``[CLS]'' token) and compute the last hidden state of ``[CLS]'', which we input to the intent scoring head, a fully connected two-layer network to yields the intent logits $x(\text{intent}=i|\mathcal{S})$ for three possible intents. The intent alignment score IAS($\mathcal{S}$) is given by the probability of the intent $i$ by applying softmax to the logits:
\begin{equation}
    \text{IAS}(\mathcal{S}) = \frac{ \exp{(x(\text{intent}=i|\mathcal{S}))} }{ \sum_{k\in \text{all intents} }\exp{(x(\text{intent}=k|\mathcal{S})) } } 
\end{equation}

The intent scorer is trained using the SciCite dataset \cite{cohan-etal-2019-structural} containing human-annotated intents. Details on training and evaluating are further described in Appendix \ref{sec:train_eval_intent_classifier}.

2. \textbf{Keyword Recall (KR)} is a metric that assesses the presence of provided keywords in the generated citation sentence. A higher value signifies that the generated sentence includes the given keywords, indicating good control over keyword incorporation. Given a generated citation sentence $\mathcal{S}$,  the keyword recall KR($\mathcal{S}$) is calculated using ROUGE-L recall to compare the keyword attribute (or a concatenated string of multiple keywords, if applicable) with $\mathcal{S}$.

3. \textbf{Fluency Score (FS)} evaluates the fluency of the generated citation sentence $\mathcal{S}$. In addition to keyword recall, the fluency metric ensures that the citation generator incorporates the keyword attributes naturally and logically without compromising fluency. Drawing inspiration from \citet{kann-etal-2018-sentence}, we employ SLOR, a language model-based fluency metric, and we normalize this score using a sigmoid function. FS($\mathcal{S}$) is calculated as follows:
\begin{equation}
    \begin{aligned}
       \text{FS}(\mathcal{S}) &= \frac{1}{ 1+ e^{- (\text{SLOR}(\mathcal{S}) - \eta)} } \\
       \text{SLOR}(\mathcal{S}) &= \frac{1}{|\mathcal{S}|}(\log(p_{LM}(\mathcal{S})) - \log(p_u(\mathcal{S})) )
    \end{aligned}
\end{equation}
In this equation, $|\mathcal{S}|$ denotes the number of tokens in sentence $\mathcal{S}$, $p_{LM}(\mathcal{S})$ represents the probability of generating the sentence with a pre-trained language model, and $p_u(\mathcal{S})$ is the product of the unconditional probabilities of all tokens in the sentence. The unconditional probability of a token refers to the probability of that token being generated as the first token in a sentence. As $\text{SLOR}(\mathcal{S})$ does not naturally range between [0,1], we apply an offset $\eta$ and a sigmoid function to normalize the score. The offset $\eta$ is introduced to the SLOR scores before sigmoid normalization, allowing greater distinguishability between fluency scores for fluent and less fluent sentences. We set the offset $\eta$ to $4$ based on empirical observations.
We utilize the 560m-Bloom \cite{workshop2023bloom} language model to calculate the SLOR because its expansive vocabulary of 250k tokens helps to avoid over-segmenting words into subtokens.

4. \textbf{ROUGE-F1} measures the textual alignment between the generated citation sentence and the ground truth. A high score is desired, as it implies that the produced sentence is informative and contextually fitting. We used ROUGE-1,2,L F1 scores also to validate the effectiveness of citation attributes in controlling generation.

\subsection{Controllability Enhancement via PPO}
\label{sec:ppo}
To enhance the controllability of our citation generator, we opted for Proximal Policy Optimization (PPO) \cite{schulman2017proximal,ramamurthy2023reinforcement} due to its capability to use controllability metrics (IAS, KR, FS, ROUGE) as rewards, facilitating a direct optimization of the LM. While cross-entropy loss during supervised fine-tuning does provide some controllability, it does not consistently ensure that specific controllability criteria are satisfied.
To illustrate, consider the ground-truth sentence centered on keywords ``\textbf{L1; sparse parameters}'': ``\textit{L1 regularization leads to sparse parameters after training.}'' Two possible generated sentences are:

A. “\textit{L1 optimization leads to dense parameters after training.}” and 

B. ``\textit{We show that L1 regularization results in sparse parameters in the model’s learned weights.}''

Despite sentence A seemingly being favored by cross-entropy loss, it misinterprets the keyword ``sparse'' and thus alters the sentence's meaning completely. Conversely, sentence B aptly captures the essence, emphasizing keyword recall. This underscores the limitation of relying solely on cross-entropy for optimal controllability. PPO's adaptability, allowing for controllability metrics as rewards, ensures the language model's outputs are both accurate and controllable, offering a more nuanced and flexible optimization strategy than other prevalent methods like adjusting beam search or 'bag of words' techniques \cite{DBLP:journals/corr/abs-2012-15416,dathathri_plug_2020}.

We used the parameters of the supervised fine-tuned model to initialize the LM. During the PPO training process, given the context and citation attributes used as input, the LM generated a batch of citation sentences, which we refer to as ``episodes''. These episodes were evaluated using the reward:
\begin{equation}
R = \frac{1}{4} \Big( \text{IAS}(\mathcal{S}) + \text{KR}(\mathcal{S}) + \text{FS}(\mathcal{S}) + \text{R}\text{S}(\mathcal{S}) \Big),
\end{equation}
where $\text{R}\text{S}(\mathcal{S})$ is the sum of ROUGE-1/2/L F1 scores, allowing its magnitude to be comparable with the other metrics.
Importance sampling was used during the optimization step, with mini-batches of episodes sampled along with their associated rewards, to estimate the expected return under the new policy using data collected with the old policy, aiming to minimize divergence from the previous policy while improving the expected return.
We implemented the PPO training using the Transformer Reinforcement Learning (TRL) framework \cite{vonwerra2022trl}.


\section{Dataset}
\label{sec:data-prep}

From a subset of arXiv computer science papers, we extracted triplets consisting of the citing paper (treated as the manuscript), the citation sentence within it, and the corresponding cited paper. The necessary components for the input context, including the local and global contexts of the manuscript, as well as the global context of the cited paper, were then extracted from these triplets.

Regarding the citation attributes, we used the SciBERT-based intent scorer, as outlined in Section \ref{sec:control-eval}, to predict the most probable citation intent for each citation sentence. To obtain the keywords attribute, we extracted noun phrases from each citation sentence and ranked them based on the cosine similarity between the keyword's Sentence-BERT (version ``all-mpnet-base-v2'', Apache License 2.0) \cite{reimers-2019-sentence-bert} embeddings and the embeddings of both the manuscript and the cited paper. This approach allowed us to retrieve up to two keywords per citation sentence.


\begin{table}
\centering
\resizebox{.9\linewidth}{!}{ 
\begin{tabular}{lccc}
\toprule
Information &  Training & Validation & Test \\
\midrule
\# samples & 233,616  &  1,299 & 1,080\\
\# citing papers & 120,425  &  1,175 & 1,005\\
\# cited papers & 69,664  &  998 & 846\\
\bottomrule
\end{tabular}
}
\caption{ \label{tab:dataset_statistics} 
The statistics of our dataset. 
}  
\end{table}

To create the training, validation, and test sets, we used a chronological split. All the citing papers used in the training set were published before March 1st, 2023, while those in the validation and test sets were published after this date. This strategy prevents any unfair advantage for the tested language models (Section \ref{sec:experiment}) by ensuring they have not previously encountered the same papers and their associated citation sentences in the validation and test sets during pretraining. 
The statistics of our dataset are shown in Table \ref{tab:dataset_statistics}.

\section{Experiment}
\label{sec:experiment}
We experimented with the encoder-decoder model BART \cite{lewis-etal-2020-bart} and decoder-only models including GPT-Neo \cite{gpt-neo}, Galactica \cite{https://doi.org/10.48550/arxiv.2211.09085}, and LLaMa \cite{touvron2023llama}, which varied in size from 125M to 7B parameters. 
All models were pretrained with corpora before March 1st, 2023.
During supervised fine-tuning, the smaller models (125M to 1.3B parameters) were fine-tuned in float16 precision with a learning rate of 1e-5. For larger models like Galactica-6.7B and LLaMa-7B, we utilized INT4 low-precision quantization \cite{dettmers2023qlora} and Low-Rank Adaptation (LoRA) \cite{hu2021lora} to reduce the GPU memory footprint. 
We set the LoRA parameters with the rank $r=16$  and the scaling factor $\alpha=32$.
For these larger models, we used a higher learning rate of 1e-4, in line with the guidance from \citet{alpaca,vonwerra2022trl}. All models were optimized with the AdamW optimizer, using default betas (0.9, 0.999), weight decay 0.05, and cosine learning rate decay. Models were fine-tuned for 5k steps with a batch size of 128 with a maximum token length of 1024. We selected the best model checkpoint based on the validation set loss every 1k steps.

To enhance controllability, we subjected Galactica and LLaMa to additional fine-tuning using Proximal Policy Optimization (PPO), as described in Section \ref{sec:ppo}. During this process, the language models generated batches of citation sentences (termed 'episodes') for subsequent mini-batch-wise backpropagation.
We set the learning rate to 1.41e-5 for all PPO fine-tuning. The PPO-finetuned models are named Galactica-125M-PPO, Galactica-6.7B-PPO, and LLaMa-7B-PPO, respectively (Table \ref{tab:performance}). More PPO hyperparameter details and hardware requirements are described in Appendix \ref{sec:ppo-train-parameters}.

In addition to fine-tuning models, we leveraged GPT-3.5-turbo-0301\footnote{\url{https://platform.openai.com/docs/models/overview}}, the backbone of ChatGPT, through prompt engineering. The aim was to compare our fine-tuned models' performance against a large language model with carefully crafted prompts. We designed specific prompts (detailed in Appendix \ref{sec:gpt-prompt}) for the ChatGPT API to generate citation sentences. We conducted this in controlled (providing context and citation attributes) and uncontrolled (providing only context) modes. 

\section{Results and Discussion}
\label{sec:res-and-diss}
We aimed to investigate several key questions: 1) whether and to what extent controllability offers advantages in citation generation, 2) whether PPO helps to improve controllability, and 3) how the model size and the nature of pretraining tasks influence overall performance. In addition, we conducted a human evaluation to compare the controllability of our fine-tuned citation generator against GPT-3.5. This evaluation provided insights beyond the automatic metrics discussed in Section \ref{sec:control-eval}.

\subsection{Controllability Is Necessary}
\begin{table}
\centering
\resizebox{\linewidth}{!}{ 
\begin{tabular}{lcccc}
\toprule
\multirow{2}*{Model} &  \multicolumn{3}{c}{ Keyword ROUGE-1 (\%) } & \multirow{2}*{ \shortstack{ Intent\\precision } } \\
\cmidrule(lr){2-4}
 & precision & recall & F1 & \\
\midrule
BART-base-140M & 22.05 & 16.70 & 17.62 & 0.6083 \\
BART-large-400M & 24.92 & 18.47 & 19.68 & \underline{0.6454} \\
GPT-Neo-125M & 21.10 & 16.36 & 17.13 & 0.5861 \\
GPT-Neo-1.3B & 28.00 & 23.18 & 23.58 & 0.6352 \\
Galactica-125M & 26.15 & 21.86 & 22.11 & 0.6204 \\
Galactica-1.3B & \textbf{29.89} & \textbf{25.53} & \textbf{25.86} & \textbf{0.6602} \\
Galactica-6.7B & \underline{29.49} & \underline{24.78} & \underline{25.10} & 0.6380 \\
LLaMa-7B & 28.13 & 22.78 & 23.40 & 0.6352 \\
\bottomrule
\end{tabular}
}
\caption{ \label{tab:res_attr_matching} 
Performance of LMs on citation attribute inference given context. 
We present ROUGE-1 metrics for keyword predictions in relation to the ground-truth keywords of the target citation sentence, alongside the precision of intent prediction, represented as the proportion of correctly inferred intents in the test set. The top scores are bolded, and the next best are underlined.
}  
\end{table}
\begin{table*}
\centering
\resizebox{\linewidth}{!}{ 
\begin{tabular}{lcccccccccccc}
\toprule
\multirow{3}*{Model} &  \multicolumn{3}{c}{ \multirow{2}*{ \shortstack{Uncontrolled\\generation} }}  & 
\multicolumn{3}{c}{ \multirow{2}*{ \shortstack{Intent-controlled\\generation} } }
& \multicolumn{6}{c}{

\multirow{2}*{ \shortstack{Intent- and keywords-controlled generation} }

} \\ \\
\cmidrule(lr){2-4}
\cmidrule(lr){5-7}
\cmidrule(lr){8-13}
& R1 & R2 & RL & R1 & R2 & RL & R1 & R2 & RL & IAS & KR & FS 
\\
\midrule
BART-base-140M & 25.49 & 4.26 & 18.28 & 26.05 & 4.52 & 18.71 & 31.63 & 8.79 & 22.74 & 0.6789 & 0.6444 & 0.7156 \\
BART-large-400M & 27.39 & 5.67 & 19.85 & 27.90 & 6.00 & 20.17 & 32.33 & 9.12 & 23.20 & 0.6521 & 0.5877 & 0.7510 \\
GPT-Neo-125M & 23.54 & 3.67 & 17.58 & 23.62 & 3.69 & 17.59 & 30.48 & 9.44 & 22.83 & 0.6252 & 0.6793 & \underline{0.7996} \\
GPT-Neo-1.3B & 28.48 & 6.12 & 20.78 & 29.04 & 6.39 & 21.28 & 36.26 & 13.48 & 26.81 & 0.7018 & 0.7936 & 0.7595 \\
Galactica-125M & 28.03 & 5.77 & 20.23 & 28.70 & 6.27 & 20.96 & 35.67 & 13.07 & 26.50 & \underline{0.7037} & 0.7914 & 0.7540 \\
Galactica-1.3B & 30.07 & \underline{7.34} & 22.06 & 30.66 & 7.62 & 22.64 & 38.06 & 15.21 & 28.50 & 0.6925 & 0.8299 & 0.7399 \\
Galactica-6.7B & \textbf{30.61} & \textbf{7.97 }& \textbf{22.59} & \underline{30.89} & \underline{8.03} & \textbf{22.87 }& \underline{38.29} & \underline{15.58} & \underline{28.70} & 0.6734 & 0.8150 & 0.7468 \\
LLaMa-7B & \underline{30.19} & 7.28 & \underline{22.13} & 30.49 & 7.46 & 22.32 & 37.71 & 14.80 & 28.30 & 0.6688 & 0.8380 & 0.7584 \\
\midrule
Galactica-125M-PPO & -- & -- & -- & 28.81 & 6.12 & 20.97 & 36.49 & 13.55 & 27.09 & \textbf{0.7273} & 0.8313 & 0.7651 \\
Galactica-6.7B-PPO & -- & -- & -- & \textbf{31.00} & \textbf{8.16} & \underline{22.85} & \textbf{38.49} & \textbf{15.81} & \textbf{28.98} & 0.6740 & 0.8334 & 0.7519 \\
LLaMa-7B-PPO & -- & -- & -- & 30.64 & 7.64 & 22.51 & 37.72 & 14.83 & 28.31 & 0.6769 & \textbf{0.8430} & 0.7591 \\


\midrule
GPT-3.5-turbo & 23.04 & 3.88 & 14.93 & 23.92 & 3.61 & 15.66 & 29.10 & 8.11 & 18.97 & 0.5716 & \underline{0.8420} & \textbf{0.8493} \\
\bottomrule
\end{tabular}
}
\caption{ \label{tab:performance}
Performance comparison of various language models (LMs) in citation generation across three operational modes: uncontrolled, intent-controlled, and intent- and keywords-controlled. The table lists ROUGE F1 scores (R1, R2, RL) in percentages, as well as Intent Alignment Score (IAS), Keyword Recall (KR), and Fluency Score (FS), with higher scores indicating superior performance. IAS, KR, and FS definitions are provided in Section \ref{sec:control-eval}.
}
\end{table*}

We assessed LMs in three modes: 1) uncontrolled mode, where the LMs utilize only the given context to infer citation attributes, and then use the inferred attributes to guide themselves 
when generating citation sentences; 2) intent-controlled mode, where we provide the gold citation intent to control generation while the keywords are still model-inferred; and 3) intent- and keywords-controlled mode, where the LMs are given all relevant input: context, citation intent, and keywords, from which the LMs generate citation sentences taking into account all available information.

Given only the context as input (same as the setting in \citet{xing_automatic_2020,ge_baco_2021}), we observed limited success of LMs in matching inferred attributes with ground-truth attributes extracted from the target citation sentence (Table \ref{tab:res_attr_matching}). Even the best-performing model, Galactica-1.3B, achieves an intent prediction precision of just 0.66 and a keyword ROUGE-1 F1 of only 0.26. 
Consequently, misinterpreted citation attributes can lead to off-topic generated citation sentences, as reflected by low ROUGE scores achieved by all LMs in the uncontrolled mode (refer to Table \ref{tab:performance}).

Intriguingly, a marked improvement in ROUGE scores was already observed when LMs generated citations merely with gold citation intent in the intent-controlled mode. Even though the citation intent is not an explicit part of the citation sentence, its presence guides LMs to generate sentences that are more aligned with the target citations, as evidenced by enhanced ROUGE scores. This enhancement is further amplified when ground-truth keywords are provided along with the citation intent, with some models witnessing a doubling in ROUGE-2 F1 scores.

\noindent\textbf{Case Study}. We further demonstrated the effectiveness of controllability with a case study in Table \ref{tab:controllability_case_study}. In the uncontrolled mode, the generated citation sentence was a background sentence and semantically mismatched with the gold citation. By just providing the citation intent ``result'', the generated sentence presented the results accurately that matched well with the gold citation. Finally, introducing the control keyword ``policy iteration-based algorithms'' further contributed to accurately including the keyword in the generated citation sentence, yielding the highest ROUGE-L F1. Our results underscore the importance of citation attribute controllability in citation generation.

\begin{table*}
  \centering
  \resizebox{\linewidth}{!}{ 
    \begin{tabularx}{\linewidth}{X}
    \toprule
 \small{ \textbf{cited title:} Reinforcement Learning: A Survey} \\
 \small{ \textbf{cited abstract:}
    This paper surveys the field of reinforcement learning from a computer-science perspective. It is written to be accessible to researchers familiar with machine learning. Both the historical basis of the field and a broad ...} \\
  \small{ \textbf{text before citation:} 
... the agent is trained against multiple environments simultaneously and gains shared experience, leading to faster convergence and improved performance. Besides, NNDP approach is value iteration–based RL algorithm, whereas our approach is policy iteration–based RL algorithm. \textbf{\textit{>>generate a citation sentence HERE<<}} } \\
    \midrule
    \\ [-1.5em]
    \small{\textbf{Ground truth citation sentence}: 
    In general, the \textcolor{lightred}{\underline{policy iteration-based algorithms}} converge \textcolor{lightred}{\underline{faster}} than \textcolor{lightred}{\underline{value iteration-based algorithms}} \#REFR, which is another reason for the superior performance of our approach. Our experimental results further support our arguments.
    }  \\
    \midrule
    \\ [-1.7em]
    \small{\textbf{Generated citation sentences of different modes:}}  \\
    \small{\textcolor{blue}{\textbf{Mode}: Uncontrolled, $\ \ $\textbf{Intent}: N/A, $\ \ $\textbf{Keywords}: N/A, $\ \ $\textbf{ROUGE-L F1}: 24.14}

     Value iteration-based RL algorithm learns the value function of each state, whereas \textcolor{lightred}{\underline{policy iteration-based}} RL algorithm learns the policy of each state \#REFR.
    } \\
    \small{\textcolor{blue}{\textbf{Mode}: Intent-controlled, $\ \ $\textbf{Intent}: result, $\ \ $\textbf{Keywords}: N/A, $\ \ $\textbf{ROUGE-L F1}: 34.62}

     It has been shown that \textcolor{lightred}{\underline{policy iteration-based}} RL algorithm converges \textcolor{lightred}{\underline{faster}} than \textcolor{lightred}{\underline{value iteration-based}} RL algorithm \#REFR.
    } \\
    \small{\textcolor{blue}{\textbf{Mode}: Intent- and keywords-controlled, \textbf{Intent}: result, \textbf{Keywords}: policy iteration-based algorithms, \textbf{ROUGE-L F1}: 50.00}

     \textcolor{lightred}{\underline{Policy iteration-based algorithms}} are known to converge \textcolor{lightred}{\underline{faster}} than \textcolor{lightred}{\underline{value iteration-based algorithms}} \#REFR.
    } \\
    \bottomrule
    \end{tabularx}%
    }
  \caption{\label{tab:controllability_case_study} 
A case study shows that citation sentences generated by Galactica-6.7B-PPO are guided by the citation intent and the keywords. 
  }
\end{table*}

\subsection{PPO Enhances LM Controllability}

We compared the LMs' performance in the intent-controlled and intent- and keywords-controlled modes before and after PPO. We excluded the uncontrolled mode from this comparison, as we assume that post-PPO, the LMs consistently operate in a controlled mode where citation attributes are given.
Comparing Galactica-125M-PPO with Galactica-125M, Galactica-6.7B-PPO with Galactica-6.7B, and LLaMa-7B-PPO with LLaMa-7B, we observed a consistent and marked improvement in the IAS, KR, FS, and ROUGE F1 scores, especially for the Galactica-6.7B-PPO model, which achieved clear improvement in all metrics and the best performance in terms of ROUGE-F1.
This result underscores the efficacy of PPO in directing the language model to adhere to the specifications of the provided attributes more effectively. Consequently, the generated citations better align with the specified intent, more comprehensively incorporate attribute keywords, and exhibit improved fluency, suggesting the effectiveness of PPO in enhancing the controllability of the LM citation generator.


\subsection{Model Size and Pretraining Matter}
A noticeable trend of improved performance accompanies the increase in language model (LM) size, underlining the role of model scale in citation generation tasks.
Additionally, notable variations in performance exist among LMs of identical sizes. Galactica consistently outperforms its counterparts, while GPT-Neo tends to underperform. Despite both models being based on the Transformer's decoder, the performance disparity can be ascribed to their distinct pretraining datasets: GPT-Neo is pretrained on a diverse corpus with only a minor portion of scientific literature, whereas Galactica's pretraining is on a large corpus of scientific texts. This result indicates that the specificity of pretraining data can significantly enhance citation generation performance, underlining the critical influence of pretraining corpora on performance.

\subsection{Human Evaluation}
\label{sec:human-eval}
\begin{table}
\centering
\resizebox{\linewidth}{!}{ 
\begin{tabular}{lccc}
\toprule
\multirow{2}*{Metric} & \multicolumn{3}{c}{ User Preference (\%) }\\
\cmidrule(lr){2-4}
& GPT-3.5-turbo & Neutral & Galactica-6.7B-PPO \\
\midrule
Intent Alignment & 17.1 & 57.1 & \textbf{25.7} \\
Keyword Recall & 15.2 & 67.6 & \textbf{17.1} \\
Fluency & \textbf{34.3} & 50.5 & 15.2\\
Similarity to GT & 24.8 & 30.5 & \textbf{44.8} \\
\bottomrule
\end{tabular}
}
\caption{Percentage distribution of user preferences for citation sentences generated by GPT-3.5-turbo and Galactica-6.7B-PPO across four criteria. ``Neutral'' indicates an equal preference for both sentences. Values in bold denote the model with a higher preference.}
\label{tab:human-eval_res}
\end{table}
Intriguing disparities emerge in our comparative analysis of GPT-3.5-turbo and the fine-tuned LM Galactica-6.7B-PPO (Table \ref{tab:performance}). Despite the second-best KR score and the best FS, GPT-3.5-turbo's performance falls short in ROUGE-F1 and IAS metrics. Such a disparity in automatic controllability metrics poses challenges in comparing the performance and controllability between models. To this end, a deeper understanding of these discrepancies was sought through a user study involving a subset of 105 examples randomly sampled from the test set. For each example, two citation sentences were presented - one generated by Galactica-6.7B-PPO, the other by GPT-3.5-turbo with the structured prompt (Appendix \ref{sec:gpt-prompt}), and the presentation order was randomized to avoid bias.
Four voluntary participants (with computer science research background, which matches well with the domain of our test dataset) were asked to express their preference using a four-criterion scale (intent alignment, keyword recall, fluency, and similarity to the ground truth), mirroring our automatic metrics (IAS, KR, FS, and ROUGE). A "no preference" option was also provided (Figure~\ref{fig:user-interface}) in case participants have equal preference for both sentences.

The results from the user study (Table \ref{tab:human-eval_res}) aligned with our automatic evaluations (Table \ref{tab:performance}). GPT-3.5-turbo-generated sentences, while preferred for fluency, often diverged significantly from the ground-truth citation sentences. This suggests that GPT-3.5-turbo struggles to generate contextually accurate citation sentences using our prompt template despite its prompting capabilities. Thus, high-capacity models like GPT-3.5-turbo may require further prompt refinement or few-shot tuning to enhance citation generation performance.








\section{Conclusion}
In this study, we introduced a controllable citation generation framework that leverages language models, highlighting the importance of user-specified attributes in the generation process. We emphasized the necessity for attribute control, underlining the complexities of citation generation, and explored the potential of enhancing controllability through Proximal Policy Optimization (PPO). Our experiments affirmed that large language models pretrained on scientific corpora are essential for citation generation, with the fine-tuned model showing advantages over GPT-3.5-turbo in both automatic metrics and human evaluations. This work provides a solid foundation for future research in controllable citation generation.

\section{Limitations}


Despite the progress achieved in this study, certain limitations are noteworthy. Due to GPU constraints, our experiments were confined to language models of up to 7B in size. While strides have been made in reducing memory footprint through low-precision quantization techniques and low-rank adaptors, fine-tuning large-scale pretrained language models, such as Galactica-120B or LLaMa-65B, on a single GPU node remains challenging. Considering the superior performance of Galactica-6.7B in citation generation, the potential performance gains of fine-tuning larger models are a promising avenue for future research.


Our study primarily concentrated on two citation attributes: citation intent and keywords, potentially overlooking other influential components. For instance, authors may cite a paper for specific details, such as certain sentences. Future work could explore conditioning citation generation systems on sentences from the cited paper, enhancing user control.


Finally, while GPT-turbo-3.5 generated citation sentences with lower ROUGE F1 scores compared to fine-tuned LMs like Galactica-6.7B-PPO, we recognize that we may not have fully harnessed the potential of these advanced large language models. For instance, enhancing the prompt with concrete generation examples could boost GPT-3.5-turbo's citation generation performance, leveraging the model's capability to learn from examples without updating parameters. This approach requires a model capable of handling long inputs, such as the "GPT-3.5-turbo-16k" with a 16k token input limit. We leave the exploration of these promising ideas to future research.

\bibliography{anthology,custom}

\begin{thebibliography}{32}
\expandafter\ifx\csname natexlab\endcsname\relax\def\natexlab#1{#1}\fi

\bibitem[{Beltagy et~al.(2019)Beltagy, Lo, and
  Cohan}]{beltagy-etal-2019-scibert}
Iz~Beltagy, Kyle Lo, and Arman Cohan. 2019.
\newblock \href {https://doi.org/10.18653/v1/D19-1371} {{S}ci{BERT}: A
  pretrained language model for scientific text}.
\newblock In \emph{Proceedings of the 2019 Conference on Empirical Methods in
  Natural Language Processing and the 9th International Joint Conference on
  Natural Language Processing (EMNLP-IJCNLP)}, pages 3615--3620, Hong Kong,
  China. Association for Computational Linguistics.

\bibitem[{Black et~al.(2021)Black, Gao, Wang, Leahy, and Biderman}]{gpt-neo}
Sid Black, Leo Gao, Phil Wang, Connor Leahy, and Stella Biderman. 2021.
\newblock \href {https://doi.org/10.5281/zenodo.5297715} {{GPT-Neo: Large Scale
  Autoregressive Language Modeling with Mesh-Tensorflow}}.
\newblock {If you use this software, please cite it using these metadata.}

\bibitem[{Chen and Zhuge(2019)}]{chen_automatic_2019}
Jingqiang Chen and Hai Zhuge. 2019.
\newblock \href {https://doi.org/10.1002/cpe.4261} {Automatic generation of
  related work through summarizing citations}.
\newblock \emph{Concurrency and Computation: Practice and Experience},
  31(3):e4261.
\newblock Number: 3 \_eprint:
  https://onlinelibrary.wiley.com/doi/pdf/10.1002/cpe.4261.

\bibitem[{Cohan et~al.(2019)Cohan, Ammar, van Zuylen, and
  Cady}]{cohan-etal-2019-structural}
Arman Cohan, Waleed Ammar, Madeleine van Zuylen, and Field Cady. 2019.
\newblock \href {https://doi.org/10.18653/v1/N19-1361} {Structural scaffolds
  for citation intent classification in scientific publications}.
\newblock In \emph{Proceedings of the 2019 Conference of the North {A}merican
  Chapter of the Association for Computational Linguistics: Human Language
  Technologies, Volume 1 (Long and Short Papers)}, pages 3586--3596,
  Minneapolis, Minnesota. Association for Computational Linguistics.

\bibitem[{Dathathri et~al.(2020)Dathathri, Madotto, Lan, Hung, Frank, Molino,
  Yosinski, and Liu}]{dathathri_plug_2020}
Sumanth Dathathri, Andrea Madotto, Janice Lan, Jane Hung, Eric Frank, Piero
  Molino, Jason Yosinski, and Rosanne Liu. 2020.
\newblock \href {http://arxiv.org/abs/1912.02164} {Plug and {Play} {Language}
  {Models}: {A} {Simple} {Approach} to {Controlled} {Text} {Generation}}.
\newblock \emph{arXiv:1912.02164 [cs]}.
\newblock ArXiv: 1912.02164.

\bibitem[{Dettmers et~al.(2023)Dettmers, Pagnoni, Holtzman, and
  Zettlemoyer}]{dettmers2023qlora}
Tim Dettmers, Artidoro Pagnoni, Ari Holtzman, and Luke Zettlemoyer. 2023.
\newblock \href {http://arxiv.org/abs/2305.14314} {Qlora: Efficient finetuning
  of quantized llms}.

\bibitem[{Ge et~al.(2021)Ge, Dinh, Liu, Su, Lu, Wang, and
  Diesner}]{ge_baco_2021}
Yubin Ge, Ly~Dinh, Xiaofeng Liu, Jinsong Su, Ziyao Lu, Ante Wang, and Jana
  Diesner. 2021.
\newblock \href {https://doi.org/10.18653/v1/2021.acl-long.116} {{BACO}: {A}
  {Background} {Knowledge}- and {Content}-{Based} {Framework} for {Citing}
  {Sentence} {Generation}}.
\newblock In \emph{Proceedings of the 59th {Annual} {Meeting} of the
  {Association} for {Computational} {Linguistics} and the 11th {International}
  {Joint} {Conference} on {Natural} {Language} {Processing} ({Volume} 1: {Long}
  {Papers})}, pages 1466--1478, Online. Association for Computational
  Linguistics.

\bibitem[{Hu et~al.(2021)Hu, Shen, Wallis, Allen-Zhu, Li, Wang, Wang, and
  Chen}]{hu2021lora}
Edward~J. Hu, Yelong Shen, Phillip Wallis, Zeyuan Allen-Zhu, Yuanzhi Li, Shean
  Wang, Lu~Wang, and Weizhu Chen. 2021.
\newblock \href {http://arxiv.org/abs/2106.09685} {Lora: Low-rank adaptation of
  large language models}.

\bibitem[{Jung et~al.(2022)Jung, Lin, Liao, Yuan, and Sun}]{jung2022intent}
Shing-Yun Jung, Ting-Han Lin, Chia-Hung Liao, Shyan-Ming Yuan, and Chuen-Tsai
  Sun. 2022.
\newblock Intent-controllable citation text generation.
\newblock \emph{Mathematics}, 10(10):1763.

\bibitem[{Jurgens et~al.(2018)Jurgens, Kumar, Hoover, McFarland, and
  Jurafsky}]{jurgens-etal-2018-measuring}
David Jurgens, Srijan Kumar, Raine Hoover, Dan McFarland, and Dan Jurafsky.
  2018.
\newblock \href {https://doi.org/10.1162/tacl_a_00028} {Measuring the evolution
  of a scientific field through citation frames}.
\newblock \emph{Transactions of the Association for Computational Linguistics},
  6:391--406.

\bibitem[{Kann et~al.(2018)Kann, Rothe, and
  Filippova}]{kann-etal-2018-sentence}
Katharina Kann, Sascha Rothe, and Katja Filippova. 2018.
\newblock \href {https://doi.org/10.18653/v1/K18-1031} {Sentence-level fluency
  evaluation: References help, but can be spared!}
\newblock In \emph{Proceedings of the 22nd Conference on Computational Natural
  Language Learning}, pages 313--323, Brussels, Belgium. Association for
  Computational Linguistics.

\bibitem[{Keskar et~al.(2019)Keskar, McCann, Varshney, Xiong, and
  Socher}]{https://doi.org/10.48550/arxiv.1909.05858}
Nitish~Shirish Keskar, Bryan McCann, Lav~R. Varshney, Caiming Xiong, and
  Richard Socher. 2019.
\newblock \href {https://doi.org/10.48550/ARXIV.1909.05858} {Ctrl: A
  conditional transformer language model for controllable generation}.

\bibitem[{Lewis et~al.(2020)Lewis, Liu, Goyal, Ghazvininejad, Mohamed, Levy,
  Stoyanov, and Zettlemoyer}]{lewis-etal-2020-bart}
Mike Lewis, Yinhan Liu, Naman Goyal, Marjan Ghazvininejad, Abdelrahman Mohamed,
  Omer Levy, Veselin Stoyanov, and Luke Zettlemoyer. 2020.
\newblock \href {https://doi.org/10.18653/v1/2020.acl-main.703} {{BART}:
  Denoising sequence-to-sequence pre-training for natural language generation,
  translation, and comprehension}.
\newblock In \emph{Proceedings of the 58th Annual Meeting of the Association
  for Computational Linguistics}, pages 7871--7880, Online. Association for
  Computational Linguistics.

\bibitem[{Lin(2004)}]{lin-2004-rouge}
Chin-Yew Lin. 2004.
\newblock \href {https://aclanthology.org/W04-1013} {{ROUGE}: A package for
  automatic evaluation of summaries}.
\newblock In \emph{Text Summarization Branches Out}, pages 74--81, Barcelona,
  Spain. Association for Computational Linguistics.

\bibitem[{Liu(2019)}]{Liu2019}
Yang Liu. 2019.
\newblock Fine-tune bert for extractive summarization.
\newblock \emph{ArXiv}.

\bibitem[{Nikiforovskaya et~al.(2020)Nikiforovskaya, Kapralov, Vlasova,
  Shpynov, and Shpilman}]{nikiforovskaya_automatic_2020}
Anna Nikiforovskaya, Nikolai Kapralov, Anna Vlasova, Oleg Shpynov, and Aleksei
  Shpilman. 2020.
\newblock \href {https://doi.org/10.1109/ICMLA51294.2020.00058} {Automatic
  generation of reviews of scientific papers}.
\newblock In \emph{2020 19th {IEEE} {International} {Conference} on {Machine}
  {Learning} and {Applications} ({ICMLA})}, pages 314--319.

\bibitem[{Pascual et~al.(2020)Pascual, Egressy, Bolli, and
  Wattenhofer}]{DBLP:journals/corr/abs-2012-15416}
Damian Pascual, Beni Egressy, Florian Bolli, and Roger Wattenhofer. 2020.
\newblock \href {http://arxiv.org/abs/2012.15416} {Directed beam search:
  Plug-and-play lexically constrained language generation}.
\newblock \emph{CoRR}, abs/2012.15416.

\bibitem[{Penders(2018)}]{2018tenrules}
Bart Penders. 2018.
\newblock \href {https://doi.org/10.1371/journal.pcbi.1006036} {Ten simple
  rules for responsible referencing}.
\newblock \emph{PLOS Computational Biology}, 14(4):e1006036.

\bibitem[{Raffel et~al.(2020)Raffel, Shazeer, Roberts, Lee, Narang, Matena,
  Zhou, Li, and Liu}]{JMLR:v21:20-074}
Colin Raffel, Noam Shazeer, Adam Roberts, Katherine Lee, Sharan Narang, Michael
  Matena, Yanqi Zhou, Wei Li, and Peter~J. Liu. 2020.
\newblock \href {http://jmlr.org/papers/v21/20-074.html} {Exploring the limits
  of transfer learning with a unified text-to-text transformer}.
\newblock \emph{Journal of Machine Learning Research}, 21(140):1--67.

\bibitem[{Ramamurthy et~al.(2023)Ramamurthy, Ammanabrolu, Brantley, Hessel,
  Sifa, Bauckhage, Hajishirzi, and Choi}]{ramamurthy2023reinforcement}
Rajkumar Ramamurthy, Prithviraj Ammanabrolu, Kianté Brantley, Jack Hessel,
  Rafet Sifa, Christian Bauckhage, Hannaneh Hajishirzi, and Yejin Choi. 2023.
\newblock \href {http://arxiv.org/abs/2210.01241} {Is reinforcement learning
  (not) for natural language processing: Benchmarks, baselines, and building
  blocks for natural language policy optimization}.

\bibitem[{Reimers and Gurevych(2019)}]{reimers-2019-sentence-bert}
Nils Reimers and Iryna Gurevych. 2019.
\newblock \href {https://arxiv.org/abs/1908.10084} {Sentence-bert: Sentence
  embeddings using siamese bert-networks}.
\newblock In \emph{Proceedings of the 2019 Conference on Empirical Methods in
  Natural Language Processing}. Association for Computational Linguistics.

\bibitem[{Schulman et~al.(2017)Schulman, Wolski, Dhariwal, Radford, and
  Klimov}]{schulman2017proximal}
John Schulman, Filip Wolski, Prafulla Dhariwal, Alec Radford, and Oleg Klimov.
  2017.
\newblock \href {http://arxiv.org/abs/1707.06347} {Proximal policy optimization
  algorithms}.

\bibitem[{Taori et~al.(2023)Taori, Gulrajani, Zhang, Dubois, Li, Guestrin,
  Liang, and Hashimoto}]{alpaca}
Rohan Taori, Ishaan Gulrajani, Tianyi Zhang, Yann Dubois, Xuechen Li, Carlos
  Guestrin, Percy Liang, and Tatsunori~B. Hashimoto. 2023.
\newblock Stanford alpaca: An instruction-following llama model.
\newblock \url{https://github.com/tatsu-lab/stanford_alpaca}.

\bibitem[{Taylor et~al.(2022)Taylor, Kardas, Cucurull, Scialom, Hartshorn,
  Saravia, Poulton, Kerkez, and
  Stojnic}]{https://doi.org/10.48550/arxiv.2211.09085}
Ross Taylor, Marcin Kardas, Guillem Cucurull, Thomas Scialom, Anthony
  Hartshorn, Elvis Saravia, Andrew Poulton, Viktor Kerkez, and Robert Stojnic.
  2022.
\newblock \href {https://doi.org/10.48550/ARXIV.2211.09085} {Galactica: A large
  language model for science}.

\bibitem[{Touvron et~al.(2023)Touvron, Lavril, Izacard, Martinet, Lachaux,
  Lacroix, Rozière, Goyal, Hambro, Azhar, Rodriguez, Joulin, Grave, and
  Lample}]{touvron2023llama}
Hugo Touvron, Thibaut Lavril, Gautier Izacard, Xavier Martinet, Marie-Anne
  Lachaux, Timothée Lacroix, Baptiste Rozière, Naman Goyal, Eric Hambro,
  Faisal Azhar, Aurelien Rodriguez, Armand Joulin, Edouard Grave, and Guillaume
  Lample. 2023.
\newblock \href {http://arxiv.org/abs/2302.13971} {Llama: Open and efficient
  foundation language models}.

\bibitem[{von Werra et~al.(2020)von Werra, Belkada, Tunstall, Beeching, Thrush,
  and Lambert}]{vonwerra2022trl}
Leandro von Werra, Younes Belkada, Lewis Tunstall, Edward Beeching, Tristan
  Thrush, and Nathan Lambert. 2020.
\newblock Trl: Transformer reinforcement learning.
\newblock \url{https://github.com/lvwerra/trl}.

\bibitem[{Wang et~al.(2022)Wang, Song, Li, Cheng, Ju, Zhang, and
  Wang}]{Wang_Song_Li_Cheng_Ju_Zhang_Wang_2022}
Yifan Wang, Yiping Song, Shuai Li, Chaoran Cheng, Wei Ju, Ming Zhang, and Sheng
  Wang. 2022.
\newblock \href {https://doi.org/10.1609/aaai.v36i10.21397} {Disencite:
  Graph-based disentangled representation learning for context-specific
  citation generation}.
\newblock \emph{Proceedings of the AAAI Conference on Artificial Intelligence},
  36(10):11449--11458.

\bibitem[{Workshop et~al.(2023)Workshop, :, Scao, Fan, Akiki, Pavlick, Ilić,
  Hesslow, Castagné, Luccioni, Yvon, Gallé, Tow, Rush, Biderman, Webson,
  Ammanamanchi, Wang, Sagot, Muennighoff, del Moral, Ruwase, Bawden, Bekman,
  McMillan-Major, Beltagy, Nguyen, Saulnier, Tan, Suarez, Sanh, Laurençon,
  Jernite, Launay, Mitchell, Raffel, Gokaslan, Simhi, Soroa, Aji, Alfassy,
  Rogers, Nitzav, Xu, Mou, Emezue, Klamm, Leong, van Strien, Adelani, Radev,
  Ponferrada, Levkovizh, Kim, Natan, Toni, Dupont, Kruszewski, Pistilli,
  Elsahar, Benyamina, Tran, Yu, Abdulmumin, Johnson, Gonzalez-Dios, de~la Rosa,
  Chim, Dodge, Zhu, Chang, Frohberg, Tobing, Bhattacharjee, Almubarak, Chen,
  Lo, Werra, Weber, Phan, allal, Tanguy, Dey, Muñoz, Masoud, Grandury,
  Šaško, Huang, Coavoux, Singh, Jiang, Vu, Jauhar, Ghaleb, Subramani,
  Kassner, Khamis, Nguyen, Espejel, de~Gibert, Villegas, Henderson, Colombo,
  Amuok, Lhoest, Harliman, Bommasani, López, Ribeiro, Osei, Pyysalo, Nagel,
  Bose, Muhammad, Sharma, Longpre, Nikpoor, Silberberg, Pai, Zink, Torrent,
  Schick, Thrush, Danchev, Nikoulina, Laippala, Lepercq, Prabhu, Alyafeai,
  Talat, Raja, Heinzerling, Si, Taşar, Salesky, Mielke, Lee, Sharma, Santilli,
  Chaffin, Stiegler, Datta, Szczechla, Chhablani, Wang, Pandey, Strobelt,
  Fries, Rozen, Gao, Sutawika, Bari, Al-shaibani, Manica, Nayak, Teehan,
  Albanie, Shen, Ben-David, Bach, Kim, Bers, Fevry, Neeraj, Thakker, Raunak,
  Tang, Yong, Sun, Brody, Uri, Tojarieh, Roberts, Chung, Tae, Phang, Press, Li,
  Narayanan, Bourfoune, Casper, Rasley, Ryabinin, Mishra, Zhang, Shoeybi,
  Peyrounette, Patry, Tazi, Sanseviero, von Platen, Cornette, Lavallée,
  Lacroix, Rajbhandari, Gandhi, Smith, Requena, Patil, Dettmers, Baruwa, Singh,
  Cheveleva, Ligozat, Subramonian, Névéol, Lovering, Garrette, Tunuguntla,
  Reiter, Taktasheva, Voloshina, Bogdanov, Winata, Schoelkopf, Kalo, Novikova,
  Forde, Clive, Kasai, Kawamura, Hazan, Carpuat, Clinciu, Kim, Cheng, Serikov,
  Antverg, van~der Wal, Zhang, Zhang, Gehrmann, Mirkin, Pais, Shavrina,
  Scialom, Yun, Limisiewicz, Rieser, Protasov, Mikhailov, Pruksachatkun,
  Belinkov, Bamberger, Kasner, Rueda, Pestana, Feizpour, Khan, Faranak, Santos,
  Hevia, Unldreaj, Aghagol, Abdollahi, Tammour, HajiHosseini, Behroozi,
  Ajibade, Saxena, Ferrandis, Contractor, Lansky, David, Kiela, Nguyen, Tan,
  Baylor, Ozoani, Mirza, Ononiwu, Rezanejad, Jones, Bhattacharya, Solaiman,
  Sedenko, Nejadgholi, Passmore, Seltzer, Sanz, Dutra, Samagaio, Elbadri,
  Mieskes, Gerchick, Akinlolu, McKenna, Qiu, Ghauri, Burynok, Abrar, Rajani,
  Elkott, Fahmy, Samuel, An, Kromann, Hao, Alizadeh, Shubber, Wang, Roy,
  Viguier, Le, Oyebade, Le, Yang, Nguyen, Kashyap, Palasciano, Callahan,
  Shukla, Miranda-Escalada, Singh, Beilharz, Wang, Brito, Zhou, Jain, Xu,
  Fourrier, Periñán, Molano, Yu, Manjavacas, Barth, Fuhrimann, Altay, Bayrak,
  Burns, Vrabec, Bello, Dash, Kang, Giorgi, Golde, Posada, Sivaraman,
  Bulchandani, Liu, Shinzato, de~Bykhovetz, Takeuchi, Pàmies, Castillo,
  Nezhurina, Sänger, Samwald, Cullan, Weinberg, Wolf, Mihaljcic, Liu,
  Freidank, Kang, Seelam, Dahlberg, Broad, Muellner, Fung, Haller,
  Chandrasekhar, Eisenberg, Martin, Canalli, Su, Su, Cahyawijaya, Garda,
  Deshmukh, Mishra, Kiblawi, Ott, Sang-aroonsiri, Kumar, Schweter, Bharati,
  Laud, Gigant, Kainuma, Kusa, Labrak, Bajaj, Venkatraman, Xu, Xu, Xu, Tan,
  Xie, Ye, Bras, Belkada, and Wolf}]{workshop2023bloom}
BigScience Workshop, :, Teven~Le Scao, Angela Fan, Christopher Akiki, Ellie
  Pavlick, Suzana Ilić, Daniel Hesslow, Roman Castagné, Alexandra~Sasha
  Luccioni, François Yvon, Matthias Gallé, Jonathan Tow, Alexander~M. Rush,
  Stella Biderman, Albert Webson, Pawan~Sasanka Ammanamanchi, Thomas Wang,
  Benoît Sagot, Niklas Muennighoff, Albert~Villanova del Moral, Olatunji
  Ruwase, Rachel Bawden, Stas Bekman, Angelina McMillan-Major, Iz~Beltagy, Huu
  Nguyen, Lucile Saulnier, Samson Tan, Pedro~Ortiz Suarez, Victor Sanh, Hugo
  Laurençon, Yacine Jernite, Julien Launay, Margaret Mitchell, Colin Raffel,
  Aaron Gokaslan, Adi Simhi, Aitor Soroa, Alham~Fikri Aji, Amit Alfassy, Anna
  Rogers, Ariel~Kreisberg Nitzav, Canwen Xu, Chenghao Mou, Chris Emezue,
  Christopher Klamm, Colin Leong, Daniel van Strien, David~Ifeoluwa Adelani,
  Dragomir Radev, Eduardo~González Ponferrada, Efrat Levkovizh, Ethan Kim,
  Eyal~Bar Natan, Francesco~De Toni, Gérard Dupont, Germán Kruszewski, Giada
  Pistilli, Hady Elsahar, Hamza Benyamina, Hieu Tran, Ian Yu, Idris Abdulmumin,
  Isaac Johnson, Itziar Gonzalez-Dios, Javier de~la Rosa, Jenny Chim, Jesse
  Dodge, Jian Zhu, Jonathan Chang, Jörg Frohberg, Joseph Tobing, Joydeep
  Bhattacharjee, Khalid Almubarak, Kimbo Chen, Kyle Lo, Leandro~Von Werra, Leon
  Weber, Long Phan, Loubna~Ben allal, Ludovic Tanguy, Manan Dey, Manuel~Romero
  Muñoz, Maraim Masoud, María Grandury, Mario Šaško, Max Huang, Maximin
  Coavoux, Mayank Singh, Mike Tian-Jian Jiang, Minh~Chien Vu, Mohammad~A.
  Jauhar, Mustafa Ghaleb, Nishant Subramani, Nora Kassner, Nurulaqilla Khamis,
  Olivier Nguyen, Omar Espejel, Ona de~Gibert, Paulo Villegas, Peter Henderson,
  Pierre Colombo, Priscilla Amuok, Quentin Lhoest, Rheza Harliman, Rishi
  Bommasani, Roberto~Luis López, Rui Ribeiro, Salomey Osei, Sampo Pyysalo,
  Sebastian Nagel, Shamik Bose, Shamsuddeen~Hassan Muhammad, Shanya Sharma,
  Shayne Longpre, Somaieh Nikpoor, Stanislav Silberberg, Suhas Pai, Sydney
  Zink, Tiago~Timponi Torrent, Timo Schick, Tristan Thrush, Valentin Danchev,
  Vassilina Nikoulina, Veronika Laippala, Violette Lepercq, Vrinda Prabhu, Zaid
  Alyafeai, Zeerak Talat, Arun Raja, Benjamin Heinzerling, Chenglei Si,
  Davut~Emre Taşar, Elizabeth Salesky, Sabrina~J. Mielke, Wilson~Y. Lee,
  Abheesht Sharma, Andrea Santilli, Antoine Chaffin, Arnaud Stiegler, Debajyoti
  Datta, Eliza Szczechla, Gunjan Chhablani, Han Wang, Harshit Pandey, Hendrik
  Strobelt, Jason~Alan Fries, Jos Rozen, Leo Gao, Lintang Sutawika, M~Saiful
  Bari, Maged~S. Al-shaibani, Matteo Manica, Nihal Nayak, Ryan Teehan, Samuel
  Albanie, Sheng Shen, Srulik Ben-David, Stephen~H. Bach, Taewoon Kim, Tali
  Bers, Thibault Fevry, Trishala Neeraj, Urmish Thakker, Vikas Raunak, Xiangru
  Tang, Zheng-Xin Yong, Zhiqing Sun, Shaked Brody, Yallow Uri, Hadar Tojarieh,
  Adam Roberts, Hyung~Won Chung, Jaesung Tae, Jason Phang, Ofir Press, Conglong
  Li, Deepak Narayanan, Hatim Bourfoune, Jared Casper, Jeff Rasley, Max
  Ryabinin, Mayank Mishra, Minjia Zhang, Mohammad Shoeybi, Myriam Peyrounette,
  Nicolas Patry, Nouamane Tazi, Omar Sanseviero, Patrick von Platen, Pierre
  Cornette, Pierre~François Lavallée, Rémi Lacroix, Samyam Rajbhandari,
  Sanchit Gandhi, Shaden Smith, Stéphane Requena, Suraj Patil, Tim Dettmers,
  Ahmed Baruwa, Amanpreet Singh, Anastasia Cheveleva, Anne-Laure Ligozat, Arjun
  Subramonian, Aurélie Névéol, Charles Lovering, Dan Garrette, Deepak
  Tunuguntla, Ehud Reiter, Ekaterina Taktasheva, Ekaterina Voloshina, Eli
  Bogdanov, Genta~Indra Winata, Hailey Schoelkopf, Jan-Christoph Kalo,
  Jekaterina Novikova, Jessica~Zosa Forde, Jordan Clive, Jungo Kasai, Ken
  Kawamura, Liam Hazan, Marine Carpuat, Miruna Clinciu, Najoung Kim, Newton
  Cheng, Oleg Serikov, Omer Antverg, Oskar van~der Wal, Rui Zhang, Ruochen
  Zhang, Sebastian Gehrmann, Shachar Mirkin, Shani Pais, Tatiana Shavrina,
  Thomas Scialom, Tian Yun, Tomasz Limisiewicz, Verena Rieser, Vitaly Protasov,
  Vladislav Mikhailov, Yada Pruksachatkun, Yonatan Belinkov, Zachary Bamberger,
  Zdeněk Kasner, Alice Rueda, Amanda Pestana, Amir Feizpour, Ammar Khan, Amy
  Faranak, Ana Santos, Anthony Hevia, Antigona Unldreaj, Arash Aghagol, Arezoo
  Abdollahi, Aycha Tammour, Azadeh HajiHosseini, Bahareh Behroozi, Benjamin
  Ajibade, Bharat Saxena, Carlos~Muñoz Ferrandis, Danish Contractor, David
  Lansky, Davis David, Douwe Kiela, Duong~A. Nguyen, Edward Tan, Emi Baylor,
  Ezinwanne Ozoani, Fatima Mirza, Frankline Ononiwu, Habib Rezanejad, Hessie
  Jones, Indrani Bhattacharya, Irene Solaiman, Irina Sedenko, Isar Nejadgholi,
  Jesse Passmore, Josh Seltzer, Julio~Bonis Sanz, Livia Dutra, Mairon Samagaio,
  Maraim Elbadri, Margot Mieskes, Marissa Gerchick, Martha Akinlolu, Michael
  McKenna, Mike Qiu, Muhammed Ghauri, Mykola Burynok, Nafis Abrar, Nazneen
  Rajani, Nour Elkott, Nour Fahmy, Olanrewaju Samuel, Ran An, Rasmus Kromann,
  Ryan Hao, Samira Alizadeh, Sarmad Shubber, Silas Wang, Sourav Roy, Sylvain
  Viguier, Thanh Le, Tobi Oyebade, Trieu Le, Yoyo Yang, Zach Nguyen,
  Abhinav~Ramesh Kashyap, Alfredo Palasciano, Alison Callahan, Anima Shukla,
  Antonio Miranda-Escalada, Ayush Singh, Benjamin Beilharz, Bo~Wang, Caio
  Brito, Chenxi Zhou, Chirag Jain, Chuxin Xu, Clémentine Fourrier,
  Daniel~León Periñán, Daniel Molano, Dian Yu, Enrique Manjavacas, Fabio
  Barth, Florian Fuhrimann, Gabriel Altay, Giyaseddin Bayrak, Gully Burns,
  Helena~U. Vrabec, Imane Bello, Ishani Dash, Jihyun Kang, John Giorgi, Jonas
  Golde, Jose~David Posada, Karthik~Rangasai Sivaraman, Lokesh Bulchandani,
  Lu~Liu, Luisa Shinzato, Madeleine~Hahn de~Bykhovetz, Maiko Takeuchi, Marc
  Pàmies, Maria~A Castillo, Marianna Nezhurina, Mario Sänger, Matthias
  Samwald, Michael Cullan, Michael Weinberg, Michiel~De Wolf, Mina Mihaljcic,
  Minna Liu, Moritz Freidank, Myungsun Kang, Natasha Seelam, Nathan Dahlberg,
  Nicholas~Michio Broad, Nikolaus Muellner, Pascale Fung, Patrick Haller, Ramya
  Chandrasekhar, Renata Eisenberg, Robert Martin, Rodrigo Canalli, Rosaline Su,
  Ruisi Su, Samuel Cahyawijaya, Samuele Garda, Shlok~S Deshmukh, Shubhanshu
  Mishra, Sid Kiblawi, Simon Ott, Sinee Sang-aroonsiri, Srishti Kumar, Stefan
  Schweter, Sushil Bharati, Tanmay Laud, Théo Gigant, Tomoya Kainuma, Wojciech
  Kusa, Yanis Labrak, Yash~Shailesh Bajaj, Yash Venkatraman, Yifan Xu, Yingxin
  Xu, Yu~Xu, Zhe Tan, Zhongli Xie, Zifan Ye, Mathilde Bras, Younes Belkada, and
  Thomas Wolf. 2023.
\newblock \href {http://arxiv.org/abs/2211.05100} {Bloom: A 176b-parameter
  open-access multilingual language model}.

\bibitem[{Wu et~al.(2021)Wu, Shieh, Hsu, and
  Chen}]{https://doi.org/10.48550/arxiv.2112.01332}
Jia-Yan Wu, Alexander Te-Wei Shieh, Shih-Ju Hsu, and Yun-Nung Chen. 2021.
\newblock \href {https://doi.org/10.48550/ARXIV.2112.01332} {Towards generating
  citation sentences for multiple references with intent control}.

\bibitem[{Xing et~al.(2020)Xing, Fan, and Wan}]{xing_automatic_2020}
Xinyu Xing, Xiaosheng Fan, and Xiaojun Wan. 2020.
\newblock \href {https://doi.org/10.18653/v1/2020.acl-main.550} {Automatic
  {Generation} of {Citation} {Texts} in {Scholarly} {Papers}: {A} {Pilot}
  {Study}}.
\newblock In \emph{Proceedings of the 58th {Annual} {Meeting} of the
  {Association} for {Computational} {Linguistics}}, pages 6181--6190, Online.
  Association for Computational Linguistics.

\bibitem[{Yang et~al.(2022)Yang, Liu, Lei, Yang, Xue, Chen, and
  Xie}]{yang2022tailor}
Kexin Yang, Dayiheng Liu, Wenqiang Lei, Baosong Yang, Mingfeng Xue, Boxing
  Chen, and Jun Xie. 2022.
\newblock \href {http://arxiv.org/abs/2204.13362} {Tailor: A prompt-based
  approach to attribute-based controlled text generation}.

\bibitem[{Yu et~al.(2022)Yu, Yu, Tong, and Jiang}]{yu2022scientific}
Mengxia Yu, Wenhao Yu, Lingbo Tong, and Meng Jiang. 2022.
\newblock Scientific comparative argument generation.

\end{thebibliography}
\bibliographystyle{acl_natbib}

\appendix

\section{
Training and Evaluation of SciBERT-based Intent Scorer
}
\label{sec:train_eval_intent_classifier}
To train the SciBERT-based intent scorer, we employ a multitask training strategy \cite{cohan-etal-2019-structural} with the main task being citation intent classification, which aims to minimize the cross-entropy loss:
\begin{equation}
    \mathcal{L} = - \log \frac{ \exp{(x_\text{intent}(i_\text{true}))} }{ \sum_{i\in \text{all intents} }\exp{(x_\text{intent}(i) ) } } 
    \label{eq:loss-cross-entropy}
\end{equation}
Additionally, we incorporate two auxiliary classification tasks (scaffolds) \cite{cohan-etal-2019-structural} to enhance the main task's performance: 1) classifying the title of the section (from 5 normalized section titles:  Introduction, Related Work, Method, Experiments, Conclusion) in which the cited sentence appears, and 2) determining the citation worthiness of the sentence. 
We utilize a separate functional head, a two-layer fully connected network for each auxiliary task, with the SciBERT-encoded "CLS" hidden states as input. The training loss is a weighted sum of the three cross-entropy losses (Equation \eqref{eq:loss-cross-entropy}), with weights of 1.0 for the main task, and 0.05 and 0.01 for the first and second auxiliary tasks, respectively. The accuracy of the intent scorer is shown in Table \ref{tab:res-intent-classification}.

\begin{table}
\centering
\resizebox{\linewidth}{!}{ 
\begin{tabular}{lcccc}
\toprule
\multirow{2}*{ \shortstack{Intent Category\\(\# samples)}}&  Background & Method & Result & Average \\ 
& (1,014) & (613) &  (260) & (Macro) \\
\midrule 

 \multirow{1}*{\citet{jurgens-etal-2018-measuring}}
&   \multirow{1}*{ 84.7 } &  \multirow{1}*{74.7} & \multirow{1}*{78.2}  & \multirow{1}*{79.2} 
\\
 \multirow{1}*{\citet{cohan-etal-2019-structural}}
&   \multirow{1}*{ 87.8 } &  \multirow{1}*{84.9} & \multirow{1}*{79.5}  & \multirow{1}*{84.0} 
\\
 \multirow{2}*{\shortstack{SciBERT+scaffolds\\(Ours)}}
&   \multirow{2}*{ \textbf{89.1} } &  \multirow{2}*{\textbf{87.1}} & \multirow{2}*{\textbf{84.0}}  & \multirow{2}*{\textbf{86.7}} 
\\\\[-0.5em]

\bottomrule
\end{tabular}
}
\caption{ \label{tab:res-intent-classification} 
The F1 scores on three citation intent categories and the average (macro) F1, tested on the SciCite dataset created by \citet{cohan-etal-2019-structural}. 
}  
\end{table}

\section{Hardware Requirements and PPO Training Hyperparameters}
\label{sec:ppo-train-parameters}
We used 4x NVIDIA A100 80 GPUs for training and a single NVIDIA RTX A6000 for inference. 
During PPO training, we adjusted the batch and mini-batch sizes according to the model size and architecture. Specifically, we used (256, 16) for Galactica-125M, (256, 4) for Galactica-6.7B, and (32, 2) for LLaMa-7B, and we ran the PPO steps until no further improvement in reward.

\section{Prompt Templates for Querying GPT-3.5-turbo API}
\label{sec:gpt-prompt}
The prompt templates utilized to query the GPT-3.5-turbo (version 0301) API are presented in this section. As illustrated in Listing \ref{code1}, the template for uncontrolled citation generation creates a citation sentence solely based on the context. In contrast, Listing \ref{code2} demonstrates the template for the controlled citation generation mode, where citation attributes are included alongside the context to guide the generation process. 
Before sending the query message to the API, we need to provide the actual content of the parameters that are listed at the beginning of the templates, such as the cited paper's title and abstract, the manuscript’s title and abstract, and the local context (the text before the target citation sentence in the manuscript), and specify the citation intent and keywords in the controlled generation mode.
During the API call to GPT-3.5-turbo, we configured the maximum token limit to 2k and set the generation temperature at 0.1.

\lstdefinestyle{mystyle}{
  backgroundcolor=\color{backcolour}, commentstyle=\color{lightblue},
  keywordstyle=\color{magenta},
  numberstyle=\tiny\color{codegray},
  stringstyle=\color{lightblue},
  basicstyle=\ttfamily\footnotesize,
  breakatwhitespace=false,         
  breaklines=true,                 
  captionpos=b,                    
  keepspaces=true,                 
  numbersep=5pt,                  
  showspaces=false,                
  showstringspaces=false,
  showtabs=false,                  
  tabsize=2
}

\lstset{style=mystyle}
\begin{lstlisting}[language=Python, float=*, label=code1,caption=The presented template was employed for querying the GPT-3.5-turbo (version 0301) API\, instructing it to generate a citation sentence given the context\, which comprises the title and abstract of the cited paper\, the title and abstract of the manuscript\, and the local context (sentences before the target citation in the manuscript). The generation was executed in an uncontrolled mode\, without the control of explicit citation attributes.]
cited_paper_title = "****Cited paper's title****"
cited_paper_abstract = "****Cited paper's abstract****"
manuscript_title = "****Manuscript's title****"
manuscript_abstract = "****Manuscript's abstract****"
manuscript_local_text_before_citation = "****The sentences before the target citation sentence in the manuscript (local context)****"

messages = [
    {
        "role": "system", 
        "content": "You are a scientific writing assistant. Your task is to infer the citation intent and relevant keywords based on the provided context, and generate a citation sentence for a given manuscript. The citation sentence should seamlessly follow the local context, reflect the inferred citation intent, and incorporate the inferred keywords."
    },
    {
        "role": "user", 
        "content": f"""
The authors need to cite the reference paper:

Title: {cited_paper_title}
Abstract: {cited_paper_abstract}

, in the manuscript with the global context:

Title: {manuscript_title}
Abstract: {manuscript_abstract}

, immediately following the provided local context:

{manuscript_local_text_before_citation}

Your task:
Please generate a citation sentence that cites the reference paper and seamlessly follows the local context. The citation sentence should implicitly reflect one of the following citation intents and incorporate relevant keywords:

1) Background: The citation provides background information or additional context about a relevant problem, concept, approach, or topic.
2) Method: The citation refers to the use of a specific method, tool, approach, or dataset from the reference paper.
3) Result: The citation compares or contrasts the results or findings of the manuscript with those in the reference paper.

Requirements:
1. Insert the citation marker "#REFR" at the position in the sentence where the reference paper should be cited. 
2. Put the citation marker "#REFR" correctly in the generated citation sentence. The citation marker should replace the entire in-text citation (e.g., authors and year of publication), should not be enclosed in any brackets, and should be placed within the sentence before the ending punctuation.

Please return only the generated citation sentence. 
            """ 
    },
]
\end{lstlisting}

\begin{lstlisting}[language=Python, float=*, label=code2,caption=The presented template was employed for querying the GPT-3.5-turbo (version 0301) API\, instructing it to generate a citation sentence given the context\, which comprises the title and abstract of the cited paper\, the title and abstract of the manuscript\, and the local context (sentences before the target citation in the manuscript). The generation was also controlled by the specified citation attributes\, including citation intent and keywords.]
cited_paper_title = "****Cited paper's title****"
cited_paper_abstract = "****Cited paper's abstract****"
manuscript_title = "****Manuscript's title****"
manuscript_abstract = "****Manuscript's abstract****"
manuscript_local_text_before_citation = "****The sentences before the target citation sentence in the manuscript (local context)****"
citation_intent = "****Specified citation intent****"
keywords = "****Specified keywords****"

messages = [
    {
        "role": "system", 
        "content": "You are a scientific writing assistant. Your task is to generate citation sentences for a given manuscript, following detailed instructions. These instructions involve taking into account the context, desired citation intent, specific keywords in your responses."
    },
    {
        "role": "user", 
        "content": f"""
The authors need to cite the reference paper:

    Title: {cited_paper_title} 
    Abstract: {cited_paper_abstract}
    
, in the manuscript with the global context:

    Title: {manuscript_title}
    Abstract: {manuscript_abstract}

, immediately following the provided local context:

    {manuscript_local_text_before_citation}
    
Your task:
Please generate one citation sentence that cites the reference paper, seamlessly follows the local context, reflects the specified citation intent, and incorporates the specified keywords.

Requirements:
1. The generated citation sentence should reflect the citation intent: {citation_intent}. Citation intents include:
    1) background: The citation provides background information or additional context about a relevant problem, concept, approach, or topic.
    2) method: The citation refers to the use of a specific method, tool, approach, or dataset from the reference paper.
    3) result: The citation compares or contrasts the results or findings of the manuscript with those in the reference paper.
2. The generated citation sentence should contain the specified keywords: {keywords}. All the provided keywords should be used. If no keywords are specified, please infer one or two keywords by yourself and generate the citation sentence based on them.
3. Insert the citation marker "#REFR" at the position in the sentence where the reference paper should be cited. 
4. Put the citation marker "#REFR" correctly in the generated citation sentence. The citation marker should replace the entire in-text citation (e.g., authors and year of publication), should not be enclosed in any brackets, and should be placed within the sentence before the ending punctuation.

Please return only the generated citation sentence. 
            """ 
    },
]
\end{lstlisting}

\section{Influence of Beam Size on Citation Generation Performance}

\begin{table*}
\centering
\resizebox{\linewidth}{!}{ 
\begin{tabular}{lcccccccccccc}
\toprule
\multirow{3}*{Model} &  \multicolumn{3}{c}{ \multirow{2}*{ \shortstack{Uncontrolled\\generation} }}  & 
\multicolumn{3}{c}{ \multirow{2}*{ \shortstack{Intent-controlled\\generation} } }
& \multicolumn{6}{c}{

\multirow{2}*{ \shortstack{Intent- and keywords-controlled generation} }

} \\ \\
\cmidrule(lr){2-4}
\cmidrule(lr){5-7}
\cmidrule(lr){8-13}
& R1 & R2 & RL & R1 & R2 & RL & R1 & R2 & RL & IAS & KR & FS 
\\
\midrule
Galactica-125M-beam1 & 27.93 & 6.00 & 20.39 & 28.67 & 6.41 & 21.09 & 35.85 & 13.44 & 26.88 & 0.7128 & 0.7667 & 0.7539 \\
Galactica-125M-beam2 & 27.26 & 5.80 & 19.61 & 28.00 & 6.26 & 20.24 & 36.00 & 13.81 & 26.68 & 0.6946 & 0.7865 & 0.7526 \\
Galactica-125M-beam4 & 27.15 & 6.01 & 19.50 & 27.69 & 6.29 & 19.98 & 35.44 & 13.59 & 26.13 & 0.6872 & 0.7656 & 0.7466 \\
Galactica-125M-beam8 & 26.47 & 6.03 & 18.87 & 26.91 & 6.38 & 19.44 & 34.99 & 13.67 & 25.91 & 0.6724 & 0.7400 & 0.7425 \\
\bottomrule
\end{tabular}
}
\caption{ \label{tab:performance-beam-size-study}
Performance of the supervised fine-tuned Galactica-125M model on the validation set, utilizing various beam sizes for inference.
}
\end{table*}

During the citation generation process with Language Models (LMs), we conducted experiments with multiple beam sizes: 1, 2, 4, and 8. The performance metrics derived from the validation set suggest that the use of a beam size of 1 yields the most consistent results. Interestingly, larger beam sizes do not enhance the performance, but negatively influence the ROUGE scores (Table \ref{tab:performance-beam-size-study}). These results could hint that larger beam sizes introduce a broader diversity in the generated text, which may affect its precision. For instance, it could impact the accurate generation of specific topic keywords and affect the control over the generation process. In a scenario where a keyword citation attribute is provided, LMs operating with a larger beam size may opt to generate a citation sentence encompassing synonyms of the keywords rather than the keyword itself, which could consequently lead to a decrease in ROUGE scores. As a result, in our experiments, we defaulted to a beam size of 1. We will delve deeper into this phenomenon in future research.

\section{Citation Generation Examples}
\label{sec:example}

\begin{figure*}
\centering
  \includegraphics[width=\linewidth]{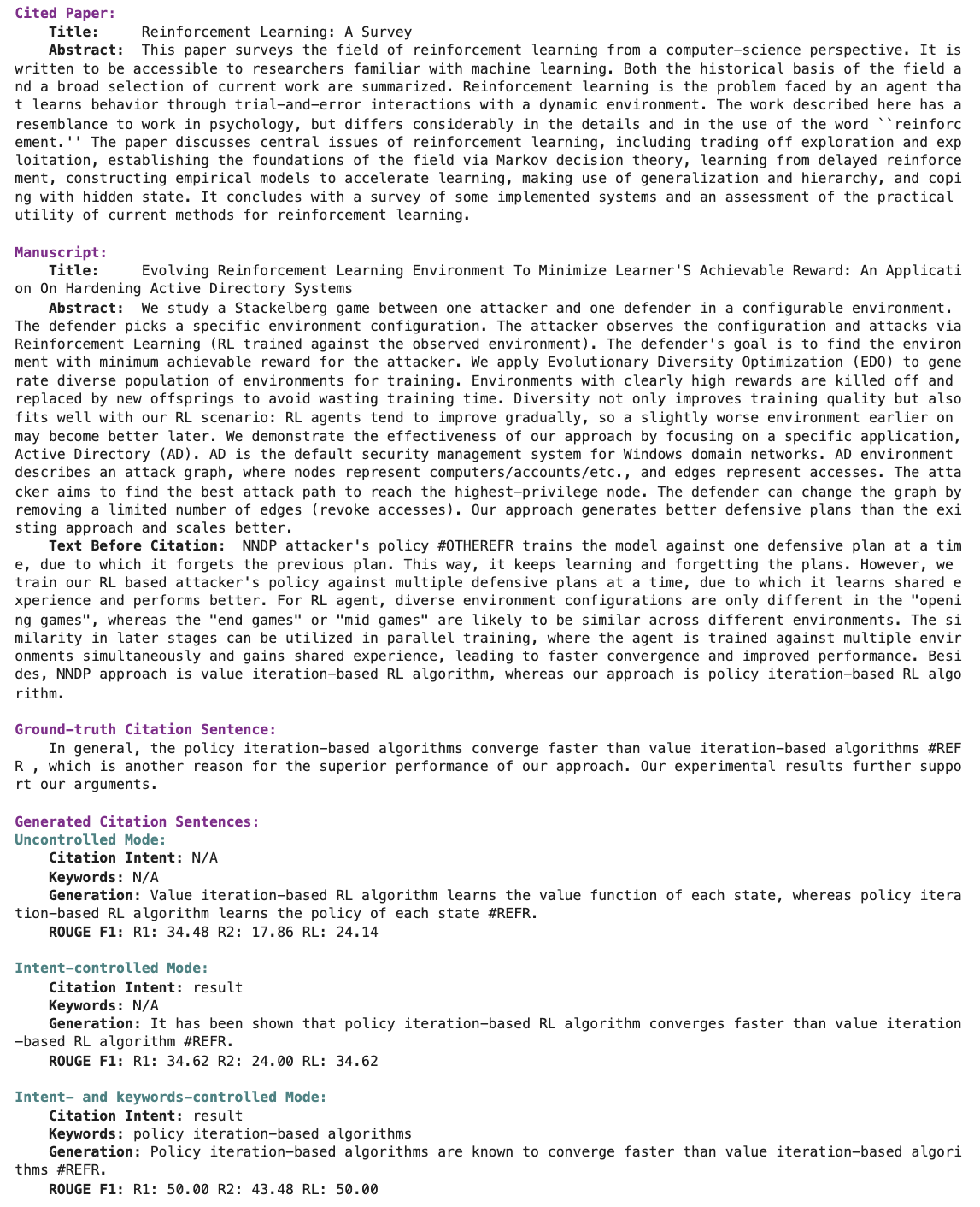}
  \caption{
Example citation sentences generated by Galactica-6.7B-PPO under the uncontrolled mode, the intent-controlled mode, and the intent- and keywords-controlled mode.
  }
  \label{fig:cit-gen-example1}
\end{figure*}

\begin{figure*}
\centering
  \includegraphics[width=\linewidth, trim=0 0 0 5, clip]{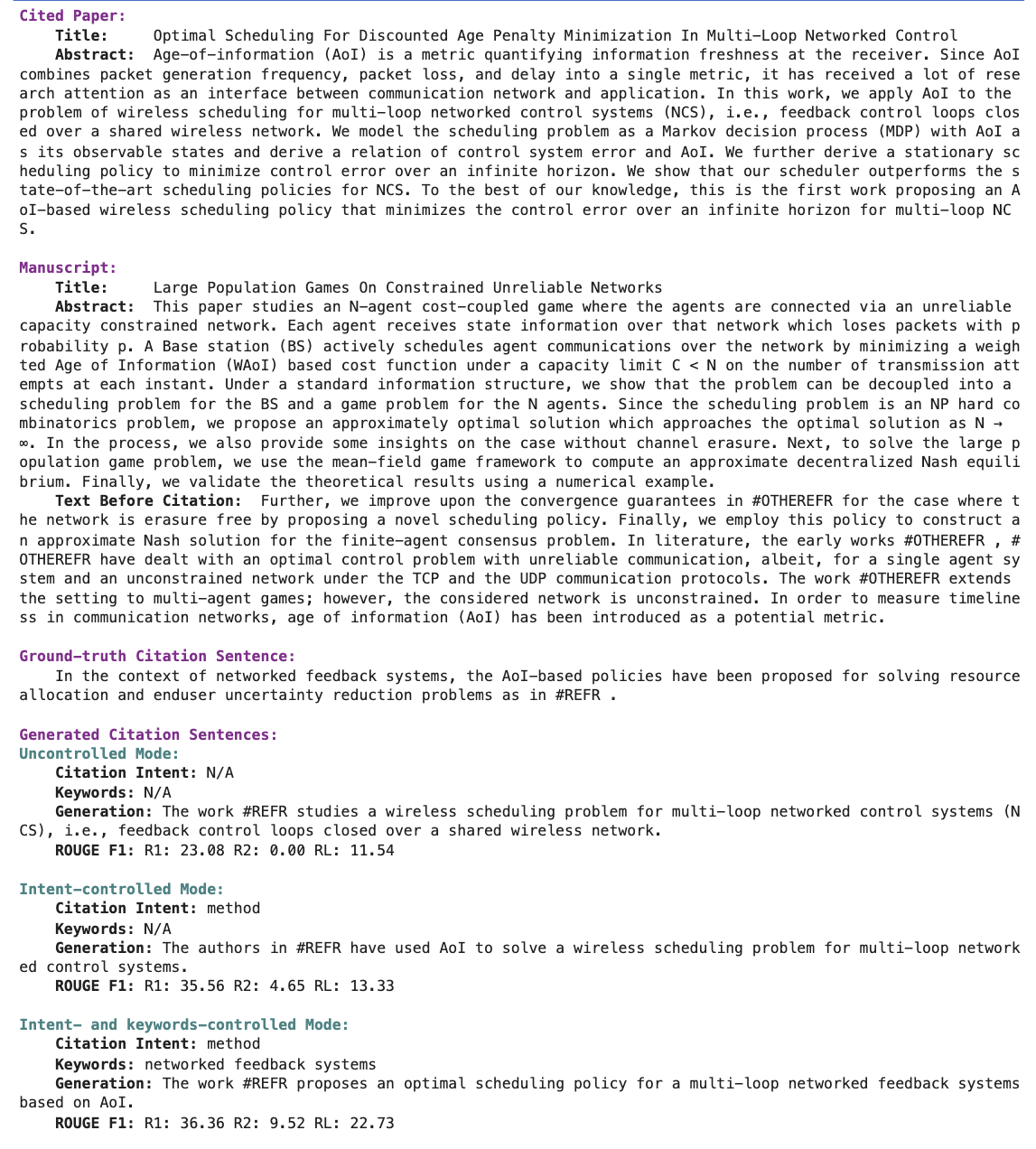}
  \caption{
Example citation sentences generated by Galactica-6.7B-PPO.
  }
  \label{fig:cit-gen-example2}
\end{figure*}

We showcase instances (Figure \ref{fig:cit-gen-example1} and \ref{fig:cit-gen-example2}) of citation sentences produced by Galactica-6.7B-PPO in three modes: 1) Uncontrolled, 2) Intent-Controlled, and 3) Intent- and Keywords-Controlled. Notably, Figure \ref{fig:cit-gen-example1} is the complete example of the case study in Table \ref{tab:controllability_case_study}. With the provision of accurate citation intent alone, the language model citation generator can align well with the stated intent, thereby enhancing ROUGE F1 scores. The addition of keyword attributes further boosts these scores.

\section{Web Interface for Human Evaluation}
\label{sec:web-interface}

\begin{figure*}
\centering
 \fbox{ \includegraphics[width=\linewidth]{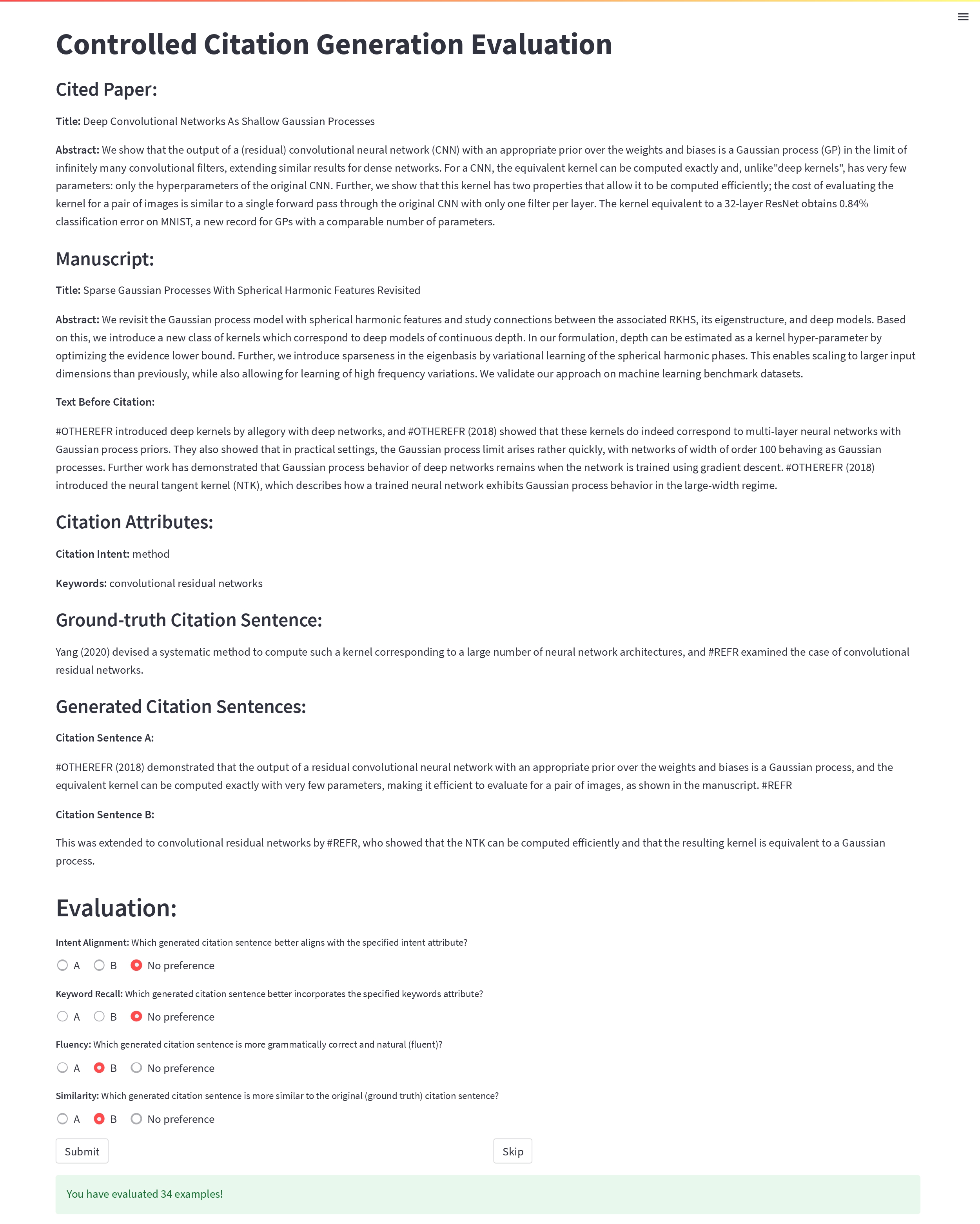}
 }
  \caption{
The user interface for the side-by-side comparison of citation sentences generated by Galactica-6.7B-PPO and GPT-3.5-turbo.
  }
  \label{fig:user-interface}
\end{figure*}

We have designed a web interface using Streamlit\footnote{\url{https://streamlit.io/}} to assess the citation sentences produced by Galactica-6.7B-PPO and GPT-3.5-turbo. As shown in Figure \ref{fig:user-interface}, this user-friendly platform enables participants to peruse context details, citation attributes, the original citation, and the model-generated sentences with ease. Upon rating the generated sentences based on the four provided criteria, participants can submit their evaluation by clicking on the ``Submit'' button, thereby preserving the data. A ``Skip'' button is also available, allowing participants to bypass any examples that fall outside their area of expertise and proceed to the next one.


\end{document}